\documentclass[letterpaper, 10 pt, conference]{ieeeconf}  

\IEEEoverridecommandlockouts                              

\usepackage{amsmath} 
\usepackage{graphicx}
\usepackage{subfigure}
\usepackage{float}
\usepackage{color}
\usepackage[utf8]{inputenc}    
\usepackage[T1]{fontenc}       

\usepackage{cite}

\title{\LARGE \bf
Building a Winning Self-Driving Car in Six Months
}

\author{Keenan Burnett, Andreas Schimpe, Sepehr Samavi, Mona Gridseth, Chengzhi Winston Liu, Qiyang Li, \\ Zachary Kroeze, and Angela P. Schoellig\thanks{The authors are with the Dynamic Systems Lab (www.dynsyslab.org) at
the University of Toronto Institute for Aerospace Studies (UTIAS), Canada.}}

\begin{document}

\bibliographystyle{IEEEtran}

\maketitle
\thispagestyle{empty}
\pagestyle{empty}

\begin{abstract}


The SAE AutoDrive Challenge is a three-year competition to develop a Level 4 autonomous vehicle by 2020. The first set of challenges were held in April of 2018 in Yuma, Arizona. Our team (aUToronto/Zeus) placed first. In this paper, we describe our complete system architecture and specialized algorithms that enabled us to win. We show that it is possible to develop a vehicle with basic autonomy features in just six months relying on simple, robust algorithms. We do not make use of a prior map. Instead, we have developed a multi-sensor visual localization solution. All of our algorithms run in real-time using CPUs only.  We also highlight the closed-loop performance of our system in detail in several experiments.

\end{abstract}


\section{Introduction}

The SAE AutoDrive Challenge \cite{sae} is a three-year competition to develop a Level 4 autonomous vehicle by 2020. The first year consisted of three individual challenges: lane-keeping through a series of tight turns, stopping at stop signs, and avoiding static objects with lane change maneuvers. These challenges were held in April of 2018 in Yuma, Arizona. Our team (aUToronto/Zeus) placed first. 

In this work, we describe our system architecture and the specialized algorithms we developed. We show that it is possible to develop a vehicle with basic autonomy features in just six months relying on simple, robust algorithms. Such an approach to address the challenges outlined above has not been shown in literature before. Our work may serve as a useful reference for researchers looking to rapidly develop their own platform to tackle open problems in self-driving. 

The majority of self-driving systems require high precision maps to navigate safely. However, mapping of the competition track prior to a scored run was prohibited and the courses did not feature multiple laps. Without a map, lane detection becomes critical for safe navigation. As a result, we developed a robust multi-sensor visual localization solution and demonstrate its closed-loop performance in significantly varying conditions.

The Intel computing platform prescribed by the competition required us to develop algorithms that would run on CPUs only. For visual perception tasks this presents a considerable challenge as most modern systems rely on GPUs to run deep neural networks \cite{autoware18, Apollo}. By adhering to this constraint, our system may prove useful to researchers looking to develop low-power autonomy.

Experimental results demonstrating closed-loop performance on self-driving cars is lacking in the literature. Hence, as a further contribution, we demonstrate the closed-loop performance of our system on each aforementioned challenge.





\begin{figure}[ht]
    \includegraphics[width=\columnwidth]{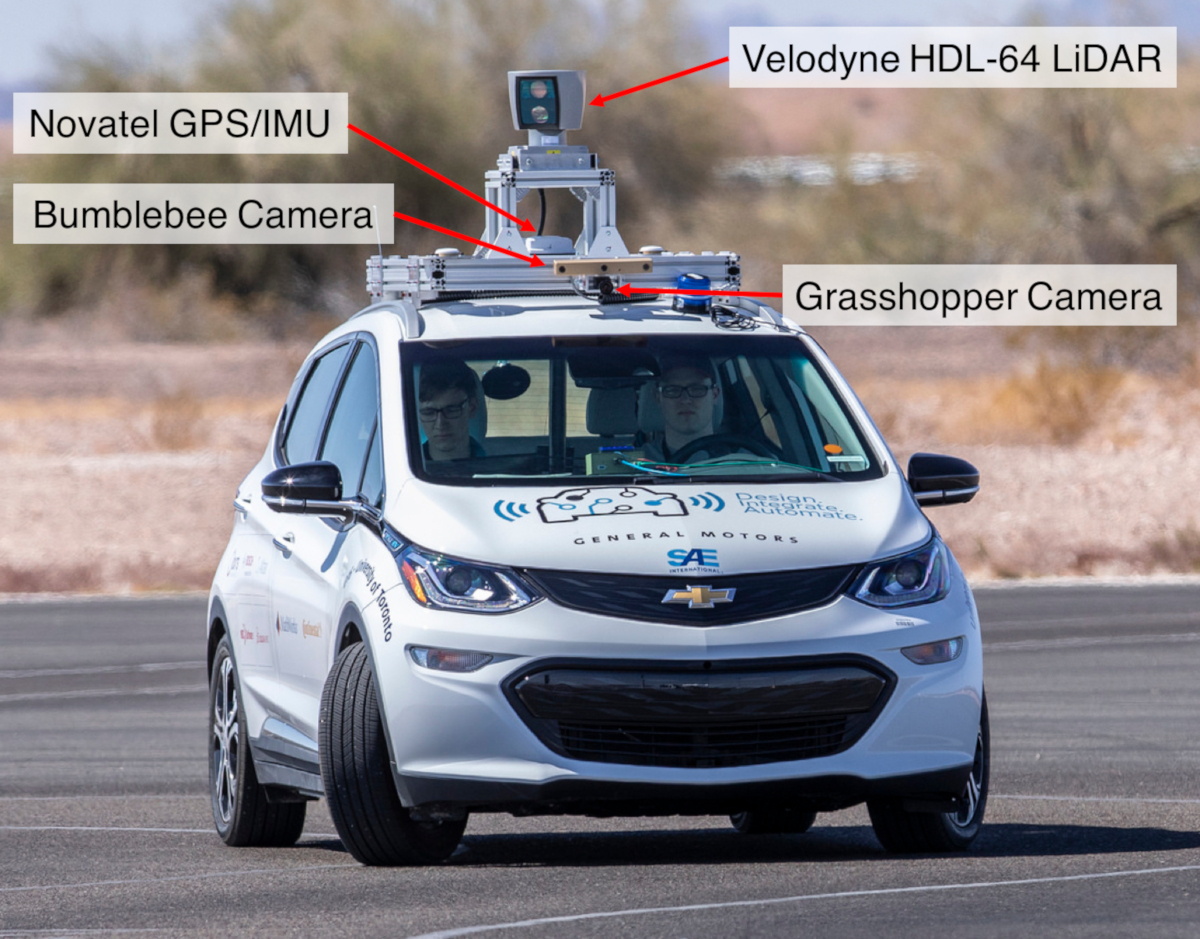}
    \centering
    \caption{Our car \textit{Zeus} at the Year 1 AutoDrive Challenge in Yuma, Arizona. Video at: http://tiny.cc/zeus-y1.} 
    \label{fig:zeus_up_close}
\end{figure}

\section{Related work}
\vspace{-1.0mm}
The DARPA Grand Challenges were the first to demonstrate the feasibility of autonomous urban driving \cite{Montemerlo2008, Urmson2008}. At a decade old, these systems are now outdated. Advances in parallel computing and computer vision are reflected in more recent systems, see \cite{Levinson2011, Wei2013}.

In 2014, the Mercedes/Bertha team navigated 100 km autonomously using only radar and vision \cite{Ziegler2014}. An update to this system for cooperative driving was presented in \cite{berthacoop}. Recent effort has also been focused on open-source self-driving. Most notable are Autoware \cite{autoware18, autoware} and Apollo \cite{Apollo}. These repositories demonstrate modern self-driving systems but rely heavily on the use of a GPU. In contrast, our software runs in real-time using CPUs only.

Popular self-driving datasets \cite{Geiger2012Kitti,Maddern2017,apolloscape} are a useful reference of sensor suites but do not provide a complete self-driving architecture as is done in this work.

Other related works include the design of an autonomous racecar \cite{fluela} and autonomously navigating rural environments \cite{autorural18}. In comparison to \cite{fluela}, we were prohibited from mapping the competition course and did not get multiple laps. Instead, we rely on a custom visual localization solution. While \cite{autorural18} estimates road boundaries for autonomous navigation, our competition track did not allow this solution. And although \cite{autorural18} does not require a detailed prior map, it still requires a topological map with GPS waypoints. This is not a requirement of our system. None of the above works present their closed-loop driving performance as is done here.



\begin{figure}[ht]
    \includegraphics[width=\columnwidth]{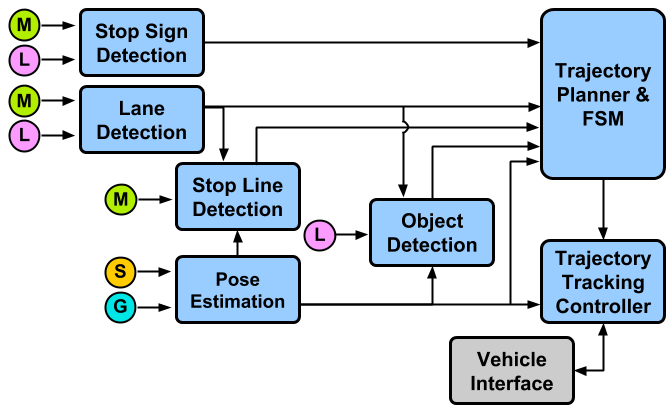}
    \centering
    \caption{Our car's software architecture. M: Monocular camera, L: 3D LiDAR, G: GPS/IMU, S: Stereo camera. The Finite State Machine (FSM) determines our current state (lane-keeping | stopping | lane changing).}
    \label{fig:sw_arch}
\end{figure}

\section{System Architecture}


Our car, \textit{Zeus}, is a 2017 Chevrolet Bolt electric vehicle. The sensors consist of a Velodyne HDL-64 3D LiDAR, a PointGrey Grasshopper monocular camera, a PointGrey Bumblebee stereo camera, and a NovAtel PwrPak7 GPS/IMU. Figure~\ref{fig:zeus_up_close} illustrates the placement of these sensors on \textit{Zeus}. Our computing platform features two Intel Xeon E5-2699R v4 CPUs. This equates to 44 physical cores running at 2.20 GHz. Our software was written in C++ using the Robot Operating System (ROS) \cite{ros}. Communication between the vehicle and our server is facilitated by a custom CAN interface developed by our team. This interface allows us to control steering, braking, propulsion, and transmission. \textit{Zeus}' software architecture is shown in Figure~\ref{fig:sw_arch}.

\section{Lane Detection}

\begin{figure}[ht]
    \includegraphics[width=\columnwidth]{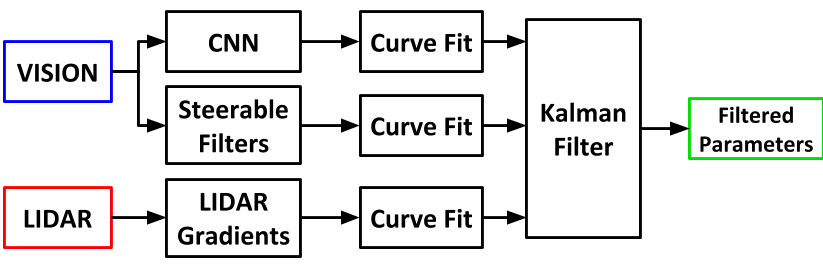}    
    \centering
    \caption{Lane detection architecture.}
    \label{fig:lane_arch}
\end{figure}

\vspace{-1mm}
Without a map, detecting lane lines is critical for safe navigation and forms the basis of our visual localization. We describe this component and its performance in detail below. This component was the most challenging to develop given our computational constraints, low-latency requirements, and unavoidable environmental variations (Yuma desert vs. Toronto winter). A diagram of our lane detection pipeline is shown in Figure~\ref{fig:lane_arch}. First, we convert incoming images into a Bird's Eye View perspective. Second, we filter the images to obtain lane marking pixels using three independent methods, two based on vision and one based on LiDAR. We then fit a centerline to each of the resulting pixel masks. This results in independent measurements of the centerline which are combined in a single Kalman filter. We model the centerline of the vehicle's ego-lane as a quadratic.


\subsection{Steerable Filters}
\vspace{-1mm}
The primary advantage of using Steerable Filters \cite{Freeman1991} is that the filters can be oriented along with the lanes. This is in contrast to approaches that make use of a set of static filters for the entire image \cite{robustlane}. These oriented filters are better at capturing lane markings at the tight corners exhibited by the Lateral Challenge. Steerable Filters can also be tuned to detect only bright lines such as lanes as opposed to all possible edges, which is important as distractors such as curbs and cracks are unavoidable in a system based on pure edge detection. The output of our steerable filters can be seen in  Figure~\ref{fig:kernel_comp}.
\vspace{-1mm}
\subsection{Convolutional Neural Network}
\begin{figure}[hb]
    \centering
    \includegraphics[width=\columnwidth]{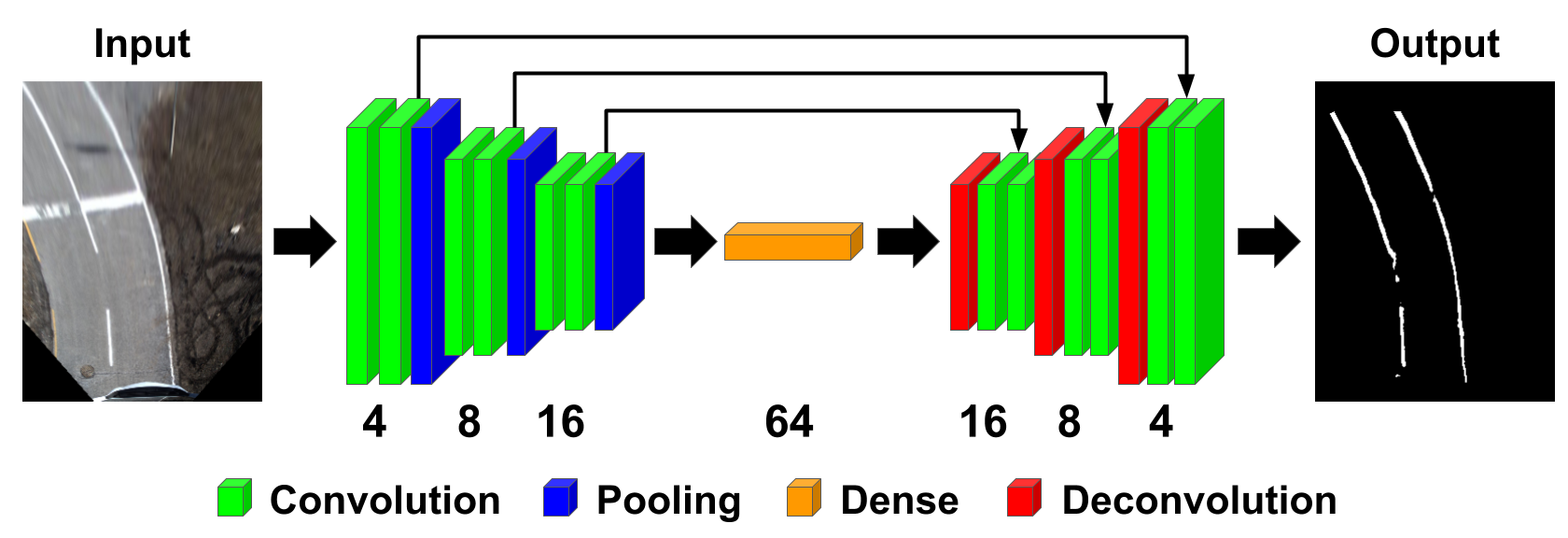}
    \caption{Convolutional Neural Network architecture. Numbers correspond to the filters. The first two convolutional layers have 4 filters each.}
    \label{fig:fcn}
\end{figure}

\vspace{-1mm}
Our team designed a Convolutional Neural Network (CNN) for the purpose of lane marking segmentation. Our network uses an encoder-decoder structure in which the encoder consists of convolution and max pooling layers and the decoder consists of deconvolution and upsampling layers. Skip connections from the encoder layers to the decoder layers allow the network to perform high-resolution upsampling as is done in \cite{Long2015}. This architecture is depicted in Figure~\ref{fig:fcn}. Given the recent success of deep learning in semantic segmentation, a deep neural network is a clear choice for this problem. However, the competition's requirement of using CPUs only represented a significant hurdle to this approach. Thus, we opted to design our own CNN with significantly less layers and filters (see Figure \ref{fig:fcn}) compared to the state-of-the-art. We trained our network in Tensorflow \cite{tensorflow} on a custom hand-labeled dataset. Random rotation, flipping, contrast, and brightness was applied to augment our dataset. The resulting network is capable of processing frames in real-time (50 Hz) on CPUs.

\subsection{LiDAR-Based Lane Detection}
\begin{figure}[h]
    \centering
    \subfigure[Bird's Eye View]{\includegraphics[width=0.42\columnwidth]{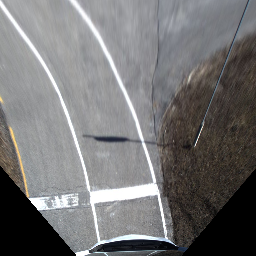}}
    \subfigure[Steerable Filters]{\includegraphics[width=0.42\columnwidth]{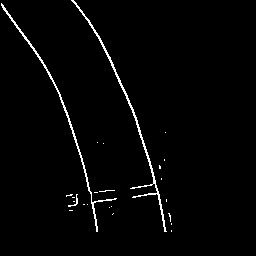}}
    \subfigure[CNN]{\includegraphics[width=0.42\columnwidth]{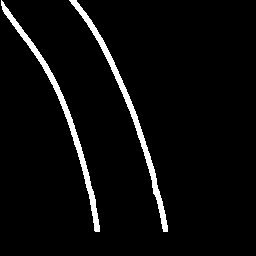}}
    \subfigure[LiDAR]{\includegraphics[width=0.42\columnwidth]{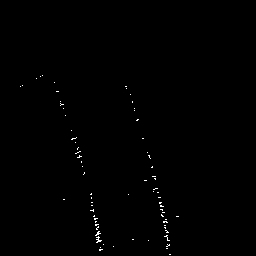}}
    \vspace{-2mm}
    \caption{Comparison between lane marking pixel extraction methods.}
    \label{fig:kernel_comp}
\end{figure}

\vspace{-1mm}
Our LiDAR-based lane detection is based on the approach in \cite{Kammel2008}. This method takes advantage of the difference in infrared reflectivity between lane marking paint and road asphalt. Each laser scan is searched for regions with high-intensity gradients. The resulting set of points are then accumulated over several laser scans in order to get a denser result. Scans are accumulated over a sliding window and aligned using Iterative Closest Point \cite{icp}. The resulting set of points are then projected into a Bird's Eye View image as seen in Figure~\ref{fig:kernel_comp}.

\vspace{10mm}

\subsection{Quadratic Curve Fitting}
The curve fitting procedure first runs RANSAC due to its capability of rejecting outliers. Linear Least Squares is then run on the inliers for a more accurate model, yielding the instantaneous parameters for our quadratic curve.

\subsection{Tracking}
This instantaneous curve fitting is not reliable enough for closed-loop driving. Due to visual distractors and noise, misdetections are not uncommon. In order to smoothly and robustly track our estimated lane parameters over time, we employ a linear Kalman filter to the lane parameters. The motion model for the lane parameters takes the following form:
\begin{equation}
	x_{k+1} = x_k + \xi_k,
\end{equation}
where $x_k = [a, b, c]^T$ is the vector of parameters corresponding to our quadratic centerline, and $\xi_k \sim \mathcal{N}(0, R)$ is zero-mean Gaussian system noise. We assume that our lane parameters stay approximately the same from one frame to the next within some variation captured by $\xi_k$. We argue that this assumption is reasonable given the relatively low speeds \textit{Zeus} operates at, and the high frame rate of our pipeline. We determined $R$ through experimentation in closed-loop driving tests. Our measurement model for each measurement is represented as:
\begin{equation}
y_k = x_k + \delta_k,
\label{meas}
\end{equation}
where $y_k$ represents the measurements from a given detection method, $\delta_k \sim \mathcal{N}(0, \Omega)$ is the measurement noise associated with that method, and
\begin{equation}
    \Omega = \begin{bmatrix}
    \sigma_a^2 & 0 & 0 \\
    0 & \sigma_b^2 & 0 \\
    0 & 0 & \sigma_c^2 \\
    \end{bmatrix}\,.
\end{equation}
    
Of course, each method has a different measurement noise. Since each detection method runs at a different frequency, we apply the measurement updates asynchronously. 

We first determined $\Omega$ for steerable filter measurements. Other measurement covariances were then chosen to be ratios of $\Omega$. We also tune the relative magnitude of system and measurement covariances, which allows a trade-off between responsiveness to measurements or smooth state estimates. Tuning this parameter was important for performance on tight corners. The entire pipeline runs in real-time (26 Hz).


\section{Stop Sign Detection}

Our Stop Sign Detector consists of a Haar Cascade classifier \cite{violajones}. Compared to an alternative deep learning object detector such as YOLO \cite{yolov2}, Haar Cascades run very fast on CPUs. They are also very simple to train and tune. We trained our classifier using OpenCV's Haar Cascade trainer on a custom hand-labeled dataset \cite{OpenCV}. We obtained the best results when images were first pre-processed using adaptive histogram equalization. We achieve 97\% precision and 90\% recall on our own challenging dataset. Our classifier runs at 26 Hz and achieves a detection range in excess of 30 m.

Once a Stop Sign is localized in a camera frame, we determine its relative depth using LiDAR. Points are first transformed into the camera frame and projected onto the image plane. We retain points that lie within the stop sign bounding box, fit a plane, and then output the distance to the plane. In order to fuse LiDAR and camera data in this manner, an accurate extrinsic calibration is required. We used an existing library for this purpose \cite{Unnikrishnan2005}. 


\section{Trajectory Planning}
The trajectory planner is comprised of a centerline planner and a lane change planner. A constant velocity profile is used if the ego-lane is unobstructed, otherwise, we generate a linearly decelerating velocity profile. The latter behaviour is also used to stop at stop lines. To change lanes, the lane change planner creates a path starting at the center of the ego-lane and ending at the center of an adjacent lane. This path is modeled as a quintic spline $f(s) = \sum_{i = 0}^5 m_i \, s^i$. To adhere to constraints on maximum lateral acceleration of the vehicle, we desire to minimize the curvature of the planned path. For this, we approximate the curvature with the concavity $\kappa(s) \approx f''(s)$ and solve the following optimization problem for the vector of coefficients $m = \begin{bmatrix} m_0 ,\, m_1 ,\, ... \,,\, m_5 \end{bmatrix}^T$:



\begin{subequations}
\begin{equation}
    \min_{m} \int_{s_0}^{s_F} \Big( f''(s) \Big)^2 \, ds 
\end{equation}
subject to
\begin{align}
    f(s_0) &= y_0 & f(s_F) &= y_F \\
    f'(s_0) &= \dot{y}_0 & f'(s_F) &= 0 \\
    f''(s_0) &= \ddot{y}_0 & f''(s_F) &= 0 \,,
\end{align}
\end{subequations}%
where $(s_0, y_0)$ and $(s_F, y_F)$ are the starting position and desired final position respectively, and $\dot{y}_0$, $\ddot{y}_0$ are the starting velocity and starting acceleration respectively.
\begin{figure}[h]
    \includegraphics[width=\columnwidth]{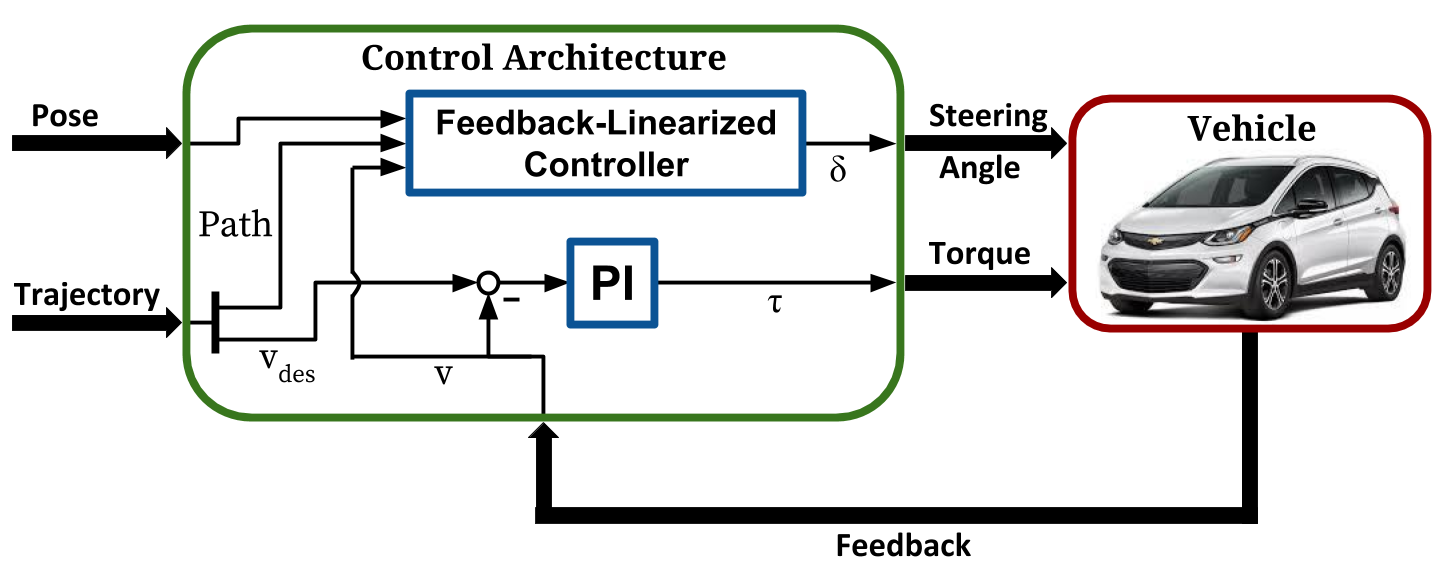}    
    \centering
    \caption{Control Architecture}
    \label{fig:control_arch}
\end{figure}

\section{Motion Control} \label{sec:motion-control}
Motion control is carried out using a decoupled control architecture as shown in Figure~\ref{fig:control_arch}. The longitudinal controller computes a torque that is commanded to the vehicle to track the desired velocity. This is done using a Proportional-Integral (PI) controller. The lateral controller is responsible for determining the steering angle to track the desired path. For this, a feedback-linearized controller (FBL)~\cite{Ostafew2013} is used.

Deviation with respect to a tracking point on the desired path, is quantified by lateral error $e^L_k$ and heading error $e^H_k$. Using the kinematic bicycle model in the rear-axle frame, their dynamics, discretized with  Forward Euler, are given by
\begin{equation}
    e_{k+1} = e_k + t_s \, \dot{e}_k = \begin{bmatrix} e^L_k \\ e^H_k \end{bmatrix} + t_s \begin{bmatrix} v \, \sin(e^H_k) \\ \frac{v}{L} \, \tan(\delta_k) \end{bmatrix},
\end{equation}
with the steering angle $\delta_k$, the vehicle wheelbase $L$ and the constant forward velocity $v$. Defining the new system states~$p_k = \begin{bmatrix} e^L_k & v \, \sin(e^H_k) \end{bmatrix}^T$, an equivalent linear system can be derived as
\begin{equation}
    p_{k+1} = p_k + t_s \, \dot{p}_k = \begin{bmatrix} 1 & t_s \\ 0 & 1 \end{bmatrix} p_k + t_s \begin{bmatrix} 0 \\ 1 \end{bmatrix} \eta_k,
\end{equation}
with the new control input $\eta_k=\frac{v^2}{L} \, \cos(e^H_k) \, \tan(\delta_k)$. Choosing a proportional controller $\eta_k = -\gamma^T p_k$ with the gains $\gamma = \begin{bmatrix} \gamma_1 & \gamma_2 \end{bmatrix}^T$, the stable closed-loop system, for~$\gamma_1,\gamma_2 > 0$, is given by
\begin{equation}
    p_{k+1} = \begin{bmatrix} 1 & t_s \\ -t_s \, \gamma_1 & 1-t_s \, \gamma_2 \end{bmatrix} p_k.
\end{equation}
Finally, with the two definitions of the new control input, the feedback-linearized control law for steering is derived as
\begin{equation}
\delta_k = \arctan \bigg(\frac{-\gamma_1 \, e^L_k - \gamma_2 \, v \, \sin(e^H_k)}{v^2 \, \cos(e^H_k)} \bigg).
\end{equation}

This control law contains the two controller gains $\gamma_1$ and $\gamma_2$. Herewith, the corrections for lateral and heading errors can be prioritized. A third parameter is the look-ahead distance along the desired path at which the tracking point is selected. This was found to have an effect on the smoothness of the steering inputs. As the Bird's Eye View image starts several meters in front of the vehicle, the centerline estimate, interpolated back to the rear-axle, was less reliable and stable. Larger look-aheads oppress some of the noise, although this results in the vehicle potentially cutting corners. 

\begin{figure}[ht]
    \centering
    \subfigure[Lane detection]{\includegraphics[height=0.4\columnwidth]{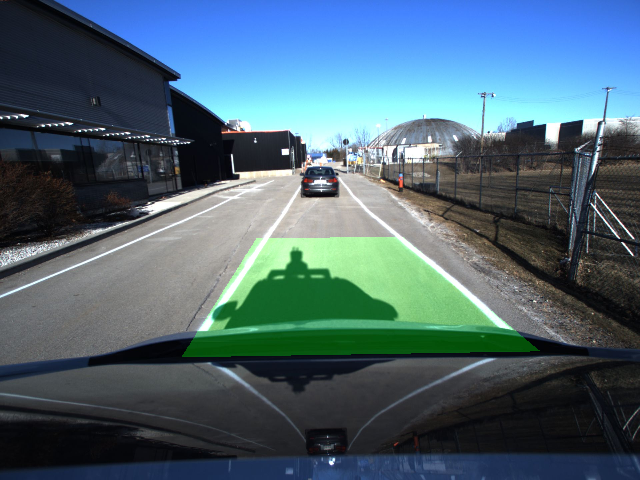}}
    \subfigure[Occupancy grid]{\includegraphics[height=0.4\columnwidth]{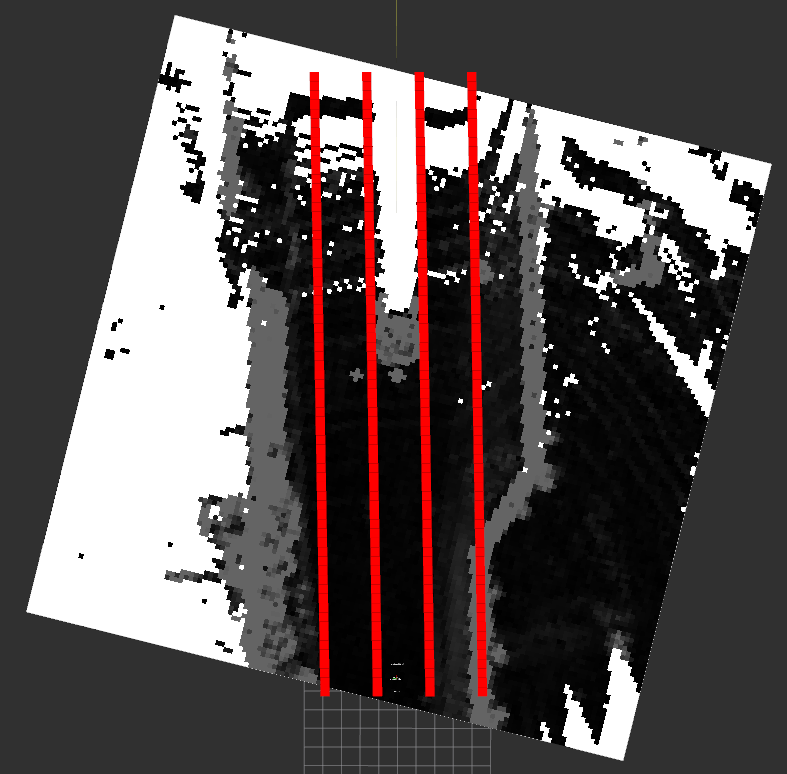}}
    \caption{Obstacle detection projects detected lane lines into an occupancy grid and looks for clusters of occupied cells (light grey).}
    \label{fig:obstacle_detection}
\end{figure}

A Model Predictive Control (MPC) approach was also developed as an alternative. The advantage of MPC is that hard constraints on lateral and longitudinal acceleration can be imposed as per competition requirements. Under the time constraints of the competition, we were unable to adequately tune the MPC to achieve better performance than the FBL controller. This is likely due to the desired reference trajectory being updated by the lane keeping module at 26 Hz. If the reference trajectory was given by a static sequence of GPS waypoints, which was not allowed by the competition, MPC would have likely been a superior choice. A comparison of MPC with FBL can be found in Section \ref{experiments}.

\section{Static Obstacle Detection}
We use the Grid Map \cite{Fankhauser2016} and Elevation Mapping \cite{Fankhauser2014} libraries to create a 2.5D gravity-aligned elevation map around the vehicle. The map covers a 35 m $\times$ 35 m area in front of the car at a resolution of 0.25 m.

Elevation alone is unreliable for detecting obstacles due to its sensitivity to pose drift. Instead, we calculate a traversability score \cite{Fankhauser2016} which is a function of map slope and roughness. We create an occupancy grid using traversability and apply smoothing to reduce spurious detections.

Given the competition's restriction to use CPUs only, it was challenging to run static occupancy mapping in real-time (10 Hz). To achieve this, we downsample incoming point clouds, and limit the horizontal field of view to $120^{\circ}$. Our map size and resolution was chosen to minimize computation without impacting our ability to detect small objects.

To detect objects, we first project detected lanes into the occupancy grid using an extrinsic transformation between our camera and LiDAR acquired using \cite{Unnikrishnan2005}. We then iterate through parts of the grid that fall between lanes lines to check for obstacles. To avoid spurious detections, we threshold on traversability and look for clusters of occupied cells. A visualization of this pipeline is shown in Figure \ref{fig:obstacle_detection}.


\begin{table*}[h]
\centering
\begin{tabular}{l|l|l|l|l|l|l|l|l|l|l|}
\cline{2-11}
                                    & \multicolumn{2}{l|}{Sunny}        & \multicolumn{2}{l|}{Shadows}      & \multicolumn{2}{l|}{Overcast}     & \multicolumn{2}{l|}{Lens Flare}   & \multicolumn{2}{l|}{Tight Corners} \\ \cline{2-11} 
                                    & RMS (m)         & F1              & RMS (m)         & F1              & RMS (m)         & F1              & RMS (m)         & F1              & RMS (m)          & F1              \\ \hline
\multicolumn{1}{|l|}{Steer}         & 0.1365          & 0.9471          & 0.2233          & 0.9433          & 0.0978          & 0.9479          & 0.2741          & 0.8849          & 0.3297           & 0.8869          \\ \hline
\multicolumn{1}{|l|}{LiDAR}         & 0.1672          & 0.9386          & 0.4172          & 0.9053          & 0.3014          & 0.8996          & 0.1649          & 0.9182          & 0.2994           & 0.7786          \\ \hline
\multicolumn{1}{|l|}{CNN}           & 0.1377          & 0.9470          & 0.2130          & 0.9487          & 0.0923          & 0.9487          & 0.3551          & 0.8655          & 0.3421           & 0.8791          \\ \hline
\multicolumn{1}{|l|}{Steer + CNN}   & \textbf{0.1344} & \textbf{0.9479} & \textbf{0.2055} & \textbf{0.9492} & 0.0976          & 0.948           & 0.1832          & 0.9101          & 0.3250           & 0.8871          \\ \hline
\multicolumn{1}{|l|}{Steer + LiDAR} & 0.1408          & 0.9466          & 0.2316          & 0.9410          & \textbf{0.0826} & \textbf{0.9504} & \textbf{0.1310} & \textbf{0.9215} & \textbf{0.2914}  & \textbf{0.9030} \\ \hline
\end{tabular}
\caption{Comparison of different lane detection approaches.}
\label{table:lane_compare_table}
\end{table*}

\section{Experiments} \label{experiments}

In this section, we describe several experiments that demonstrate the closed-loop performance of our system on tasks required by the AutoDrive Challenge. All data-collection and experiments were performed on private roads around our institute's facility. 

\begin{figure}[ht]
    \includegraphics[width=\columnwidth]{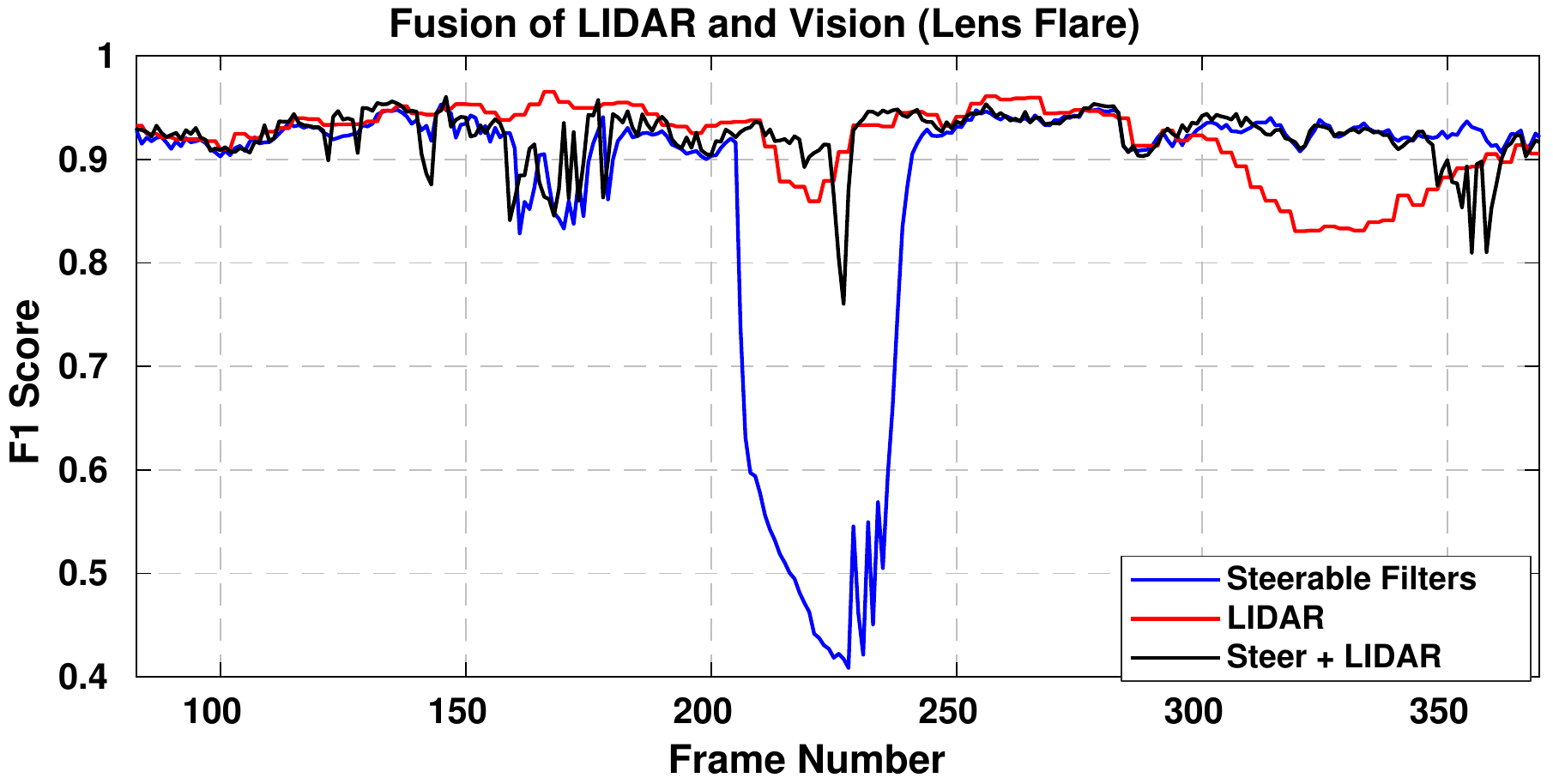}    
    \centering
    \caption{Fusion of Steerable Filters and LiDAR (Lens Flare Scenario).}
    \label{fig:steer_LiDAR}
\end{figure}

Images are first converted to a Bird's Eye View and lane markings are hand-labeled. We then fit a polygon to delimit the ground truth and detected lane markings. We compare these polygons to each other to calculate an F1 score for per-pixel classification on each frame \cite{kitti_road}. We also calculate an RMS error based on the lateral difference between our hand-labeled and detected centerline. We play back recorded data and use the output of our filtered lane detection parameters to obtain results. The results reported in Table \ref{table:lane_compare_table}.

\textbf{Sunny:} All approaches have comparable performance with vision slightly outperforming LiDAR. \textbf{Shadows:} Vision-based approaches perform worse due to overhanging shadows. LiDAR-based approach performed worse due to worse road quality on this course. \textbf{Overcast:} Vision-based algorithms improve in overcast conditions due to diffuse lighting. Our LiDAR-based approach dropped in performance possibly due to a wet road. \textbf{Lens Flare:} Vision-based algorithms suffer in the event of a lens flare, but the LiDAR-based approach performed similarly to Sunny. \textbf{Tight Corners:} High-curvature sections are difficult for all approaches.

\begin{figure}[ht]
    \centering
    \subfigure[Sunny]{\includegraphics[width=0.45\columnwidth]{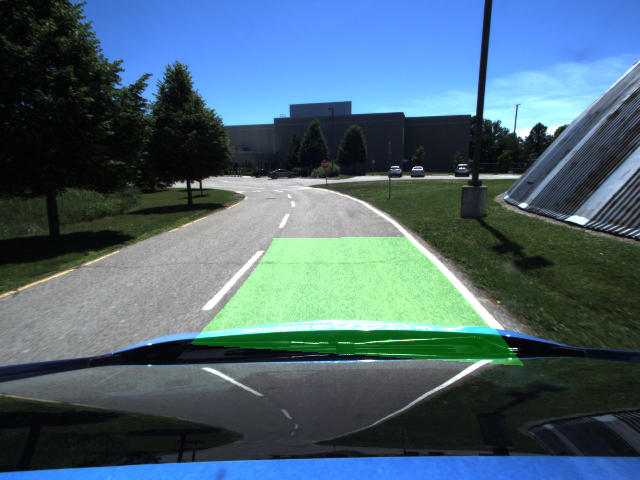}}
    \subfigure[Shadows]{\includegraphics[width=0.45\columnwidth]{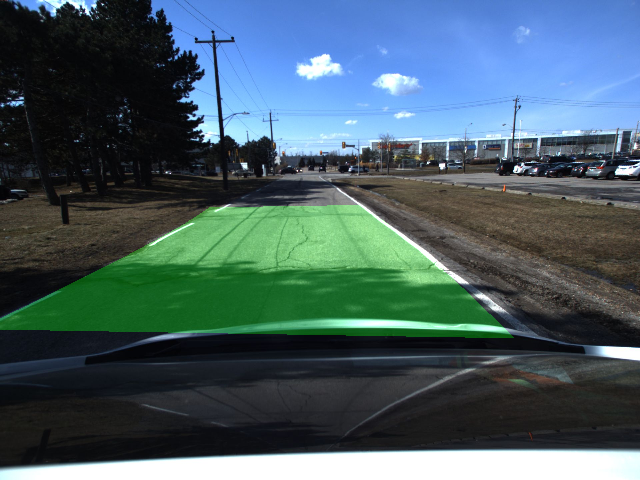}}
    \subfigure[Lens Flare]{\includegraphics[width=0.45\columnwidth]{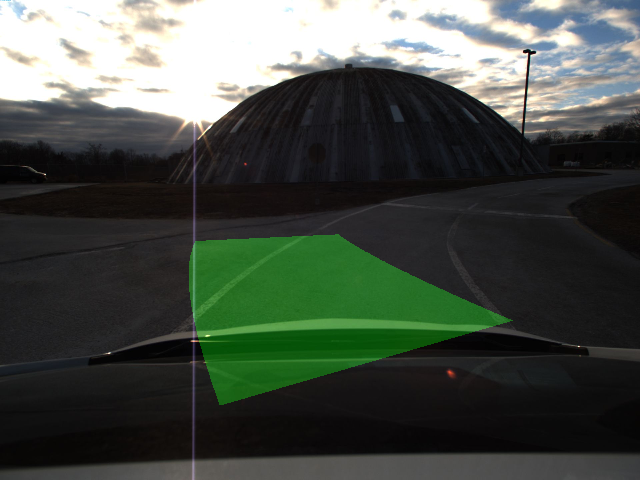}}
    \subfigure[Tight Corners]{\includegraphics[width=0.45\columnwidth]{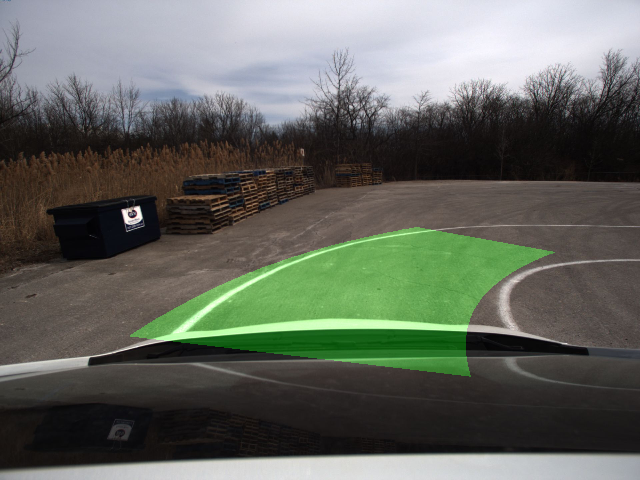}}
    \caption{Qualitative performance of Steerable Filters for lane detection.}
    \label{fig:lane_comp}
\end{figure}

These results demonstrate that the fusion of multiple approaches yields comparable, if not better performance. We have observed that when one method performs poorly for a single frame, the second method will often perform much better. In this way, the combination of multiple methods results in robust performance as shown in Figure \ref{fig:steer_LiDAR}. In the case of Shadows, a system can benefit from fusing multiple independent vision approaches. In the case of Lens Flares, it is clear that a system can benefit from the fusion of multiple sensing modalities. This benefit is also shown in Figure \ref{fig:steer_LiDAR}. At the competition, we opted to simply use Steerable filters due to the simplicity of this approach and limited development time. Qualitative performance is shown in Figure \ref{fig:lane_comp}.





\subsection{Closed-Loop Lane Keeping}

\begin{figure}[ht]
    \includegraphics[width=\columnwidth]{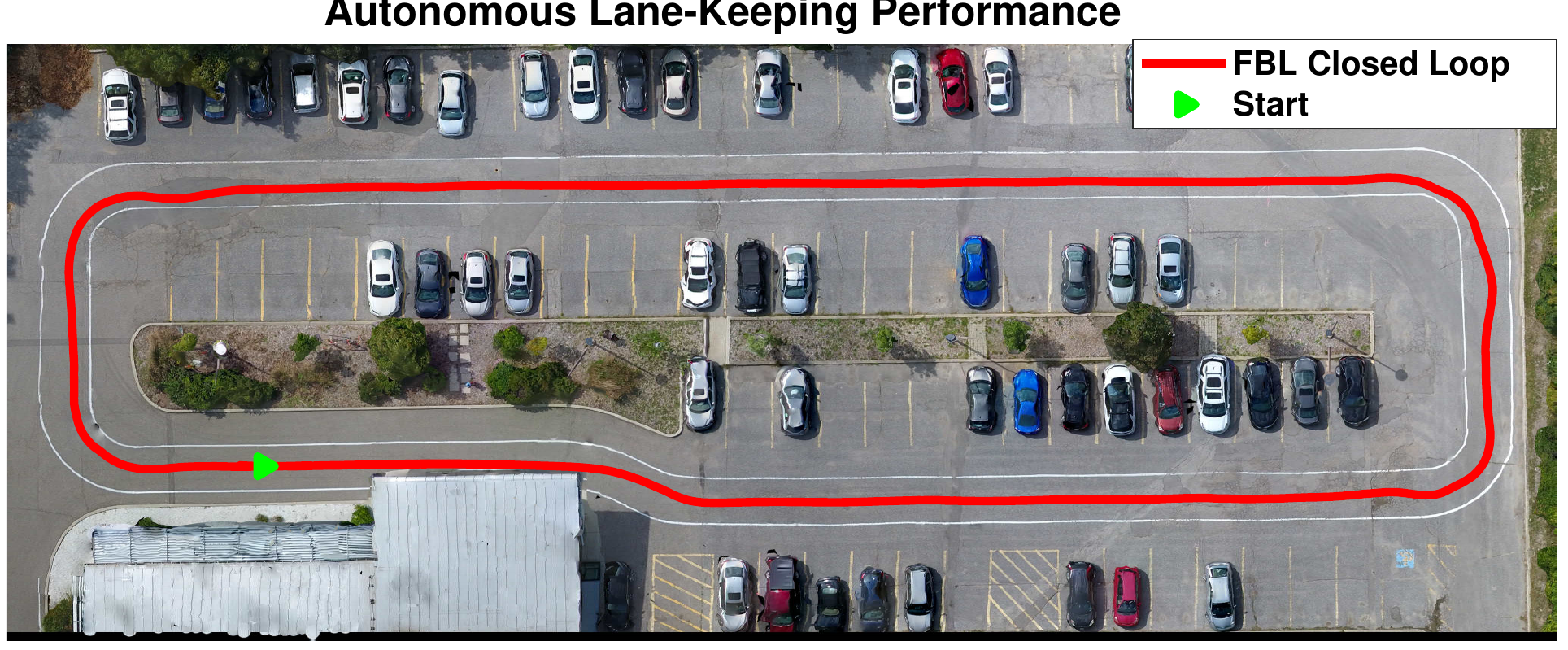}    
    \centering
    \caption{Closed-loop lane-keeping performance around our track at UTIAS.}
    \label{fig:lateral_bev}
\end{figure}

In order to benchmark our closed-loop lane-keeping performance, we compare against ground truth generated from the average of several traverses by a human driver. GPS waypoints are collected at 5 cm intervals using our Novatel PwrPak7-E1 GNSS Intertial Naviation System (INS). TerraStar corrections sent to the INS yield an RMS position accuracy of 6 cm. It is important to note that this accuracy applies for both the ground truth collection as well as the data collected for closed-loop runs. The test course is illustrated in Figure \ref{fig:lateral_bev}. The course is roughly 200 m in length, with four tight turns with a 3.0 m radius. The lanes are 3.0 m in width. We tracked a constant velocity of 2.5 m/s for this experiment. Note the lack of curbs and presence of vehicles.

\begin{figure}[ht]
    \includegraphics[width=\columnwidth]{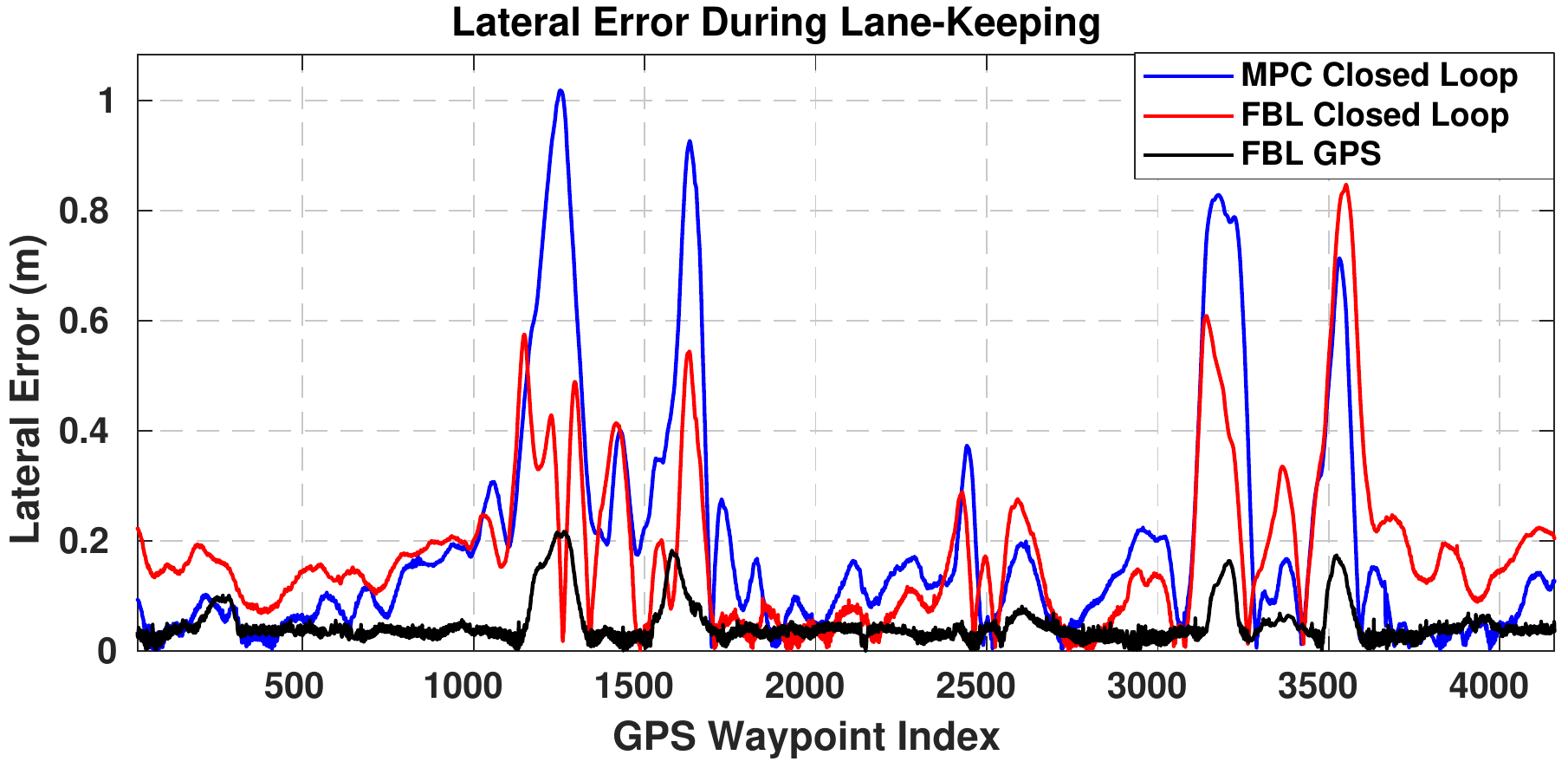}    
    \centering
    \caption{Lateral tracking error during lane-keeping.}
    \label{fig:lateral_error}
\end{figure}

In Figure \ref{fig:lateral_error}, lateral tracking error for closed-loop runs using both the FBL controller and MPC is plotted as a function of the GPS waypoint index of the manual run. Of note is that FBL and MPC have similar maximum error, but the RMS of FBL is lower, 23 cm and 29 cm, respectively. As mentioned in Section \ref{sec:motion-control}, while it would be expected that MPC would have better performance, the reference path is constantly being updated by Lane Detection at 26 Hz, and MPC has significantly more parameters to tune compared to FBL. Due to these reasons and the time constraints of the competition, we decided to use FBL for path tracking. Figure \ref{fig:lateral_error} demonstrates that we are able to achieve an RMS tracking error less than 20 cm in straight sections and less than 1 m in tight turns. This figure also shows the significant improvement in lateral error if the ground truth GPS waypoints are used as a reference, achieving an RMS error of less than 10 cm.

Due to the simplicity and the built-in redundancy of our approach, the closed-loop lane-keeping performance is highly reliable. At our competition in the desert in Yuma, Arizona, the system performed as expected with little tuning required despite the drastic difference in environmental conditions compared to winter weather in Toronto, Canada.

\subsection{Stopping at stop signs}

\begin{figure}[ht]
    \includegraphics[width=\columnwidth]{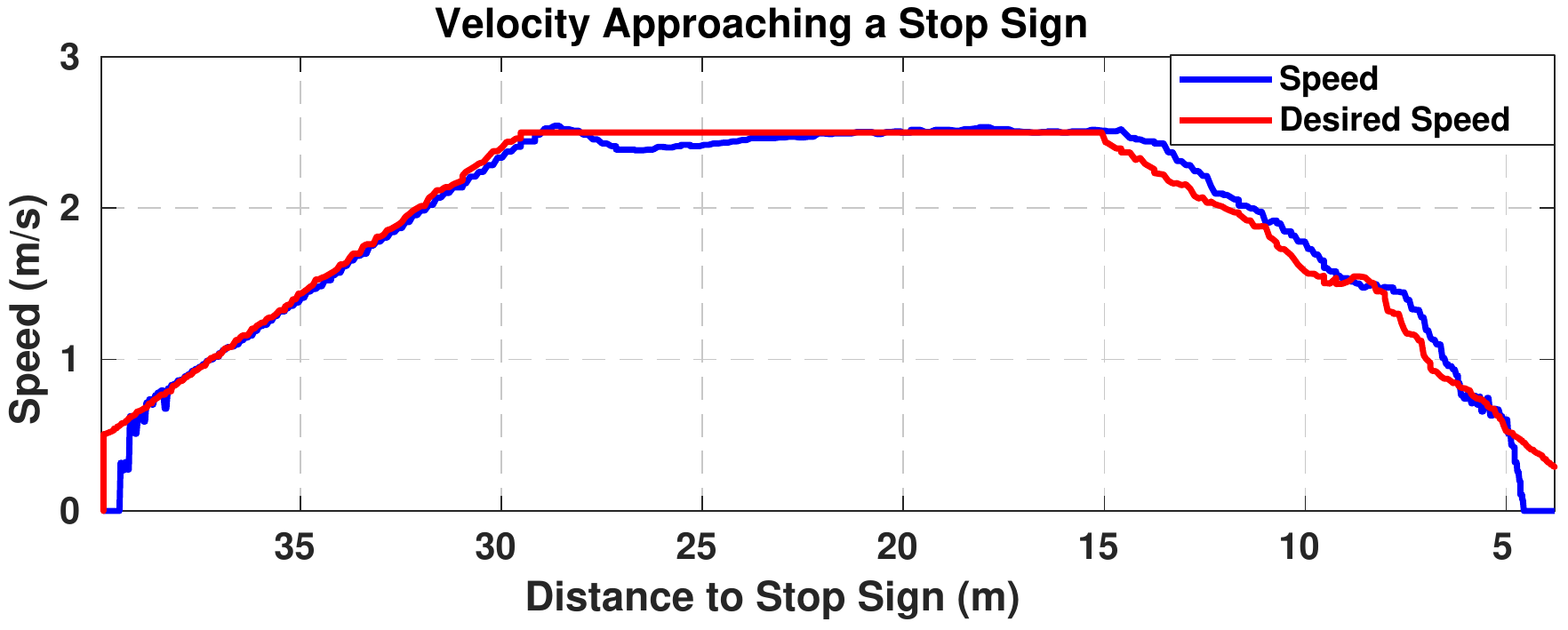}    
    \centering
    \caption{Velocity profile of \textit{Zeus} approaching a stop sign. While we are able to detect the stop sign at 30 m, we only begin slowing down depending on the desired maximum deceleration.}
    \label{fig:velprof_stop}
\end{figure}

For this experiment, we recorded the position of our stop sign and stop line in constant Universal Transverse Mercator (UTM) coordinates. In this way, we are able to extrapolate a ground truth distance from the stop sign to our current position by using the INS. In this experiment, we start from rest, accelerate to 2.5 m/s and then decelerate to a stop at the stop sign. Figure \ref{fig:velprof_stop} demonstrates the performance of our velocity controller as we approach a stop sign. This experiment was repeated 10 times and achieved an average stop distance of 14 cm, maximum stop distance of 29 cm, and a minimum stop distance of 1 cm to the stop line.

 \begin{figure}[ht]
    \includegraphics[width=\columnwidth]{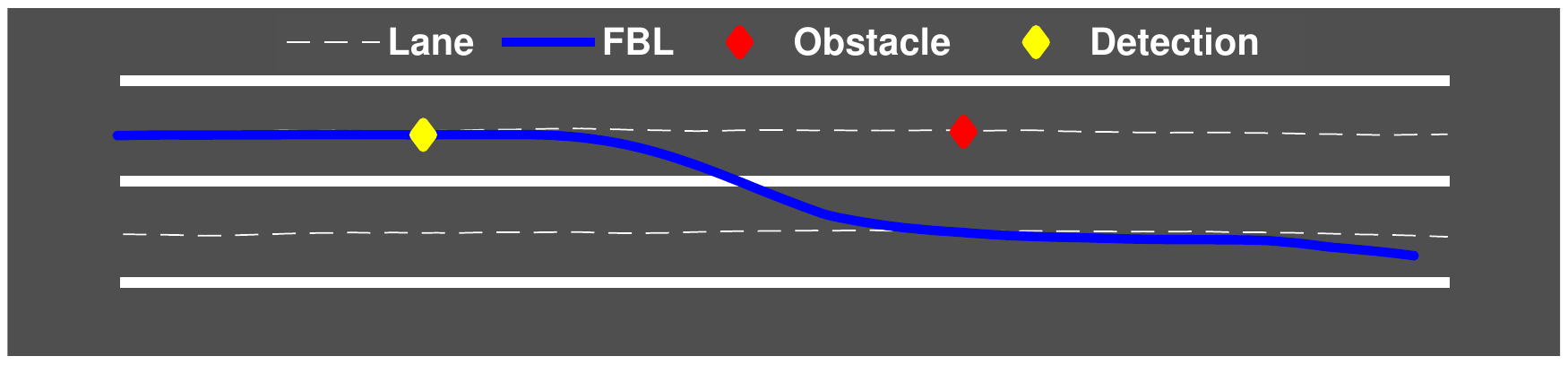}    
    \centering
    \caption{Closed-loop lane changing in the presence of a static obstruction in the ego-lane. The closed-loop path is shown in blue. We are able to detect obstacles at 30 m and plan an appropriate lane change maneuver that respects the lateral acceleration constraints imposed by the competition.}
    \label{fig:oda}
\end{figure}

\subsection{Obstacle Avoidance}
For this experiment, we first recorded the location of the obstacle and obtained the location of the lane centers in UTM coordinates. We use Lane Detection to generate the lane center of the current lane, as well as available adjacent lanes, until a lane change is desired. At this point, we construct a quintic spline between the measured lane centers to obtain ground truth. Figure \ref{fig:oda} illustrates the test course as well as the paths that our system is able to generate. Note the smoothness of the lane change trajectory and the high quality lane keeping performance before and after the lane change maneuver. In this experiment, we detect the obstacle 24 m away. We are also able to detect varying sizes of objects, down to a 1-meter tall and 0.75-meter wide pylon.

\section{Conclusions}

In this paper, we described our approach to the Year 1 SAE AutoDrive Challenge. We demonstrated a multi-modal visual localization solution that formed the basis of our autonomy suite. We demonstrated the performance of our system in several closed-loop experiments. By focusing on simple approaches with added redundancy, we were able to build a system sufficiently reliable to win the competition in six months. In fact, very few tuning parameter or code changes were required once at the competition and the system operated largely as expected. Year 2 of the SAE AutoDrive Challenge is to be held at MCity, Michigan in June 2019.


 Map-less solutions result in high reliability and are less susceptible to GPS failures. Nevertheless, our future work will include developing a mapping suite to encode the location of lanes, signs, and intersections. In addition, a more comprehensive planning system will be required to handle intersections and traffic lights. We will be looking to augment our pose estimation system, possibly with a vision- or LIDAR-based SLAM solution. Our constraint to use CPUs only will be removed. As such, we will be looking to jump-start our team's development using powerful deep learning tools for all visual perception tasks. We will be looking to accelerate these networks on a combination of CPUs and FPGAs.


\newpage

\bibliography{IEEEabrv,bib/refs}

\end{document}